\pdfoutput=1

\documentclass[11pt]{article}

\usepackage[]{acl}

\usepackage{times}
\usepackage{latexsym}

\usepackage[T1]{fontenc}

\usepackage[utf8]{inputenc}

\usepackage{microtype}

\usepackage{inconsolata}

\usepackage{graphicx} 
\usepackage{caption} 
\usepackage{algorithm}
\usepackage{algorithmic}

\usepackage{hyperref} 

\usepackage{booktabs} 
\usepackage{multirow}
\usepackage{tabularx}
\usepackage{amsmath,amsfonts,amssymb}

\usepackage{subcaption}

\usepackage{mathtools}

\usepackage{enumitem}

\usepackage{nicefrac}       

\title{A Classification-Guided Approach for Adversarial Attacks\\ against Neural Machine Translation}

\author{Sahar Sarizadeh \\
  EPFL, Switzerland \\
  \texttt{\small sahar.sadrizadeh@epfl.ch} \\\And
  Ljiljana Dolamic \\
  Armasuisse S+T, Switzerland \\
  \texttt{\small ljiljana.dolamic@ar.admin.ch} \\ \And
  Pascal Frossard \\
  EPFL, Switzerland \\
  \texttt{\small pascal.frossard@epfl.ch} \\
  }

\begin{document}
\maketitle
\begin{abstract}
\looseness=-1 Neural Machine Translation (NMT) models have been shown to be vulnerable to adversarial attacks, wherein carefully crafted perturbations of the input can mislead the target model. In this paper, we introduce ACT, a novel adversarial attack framework against NMT systems guided by a classifier. In our attack, the adversary aims to craft meaning-preserving adversarial examples whose translations in the target language by the NMT model belong to a different class than the original translations. Unlike previous attacks, our new approach has a more substantial effect on the translation by altering the overall meaning, which then leads to a different class determined by an oracle classifier. To evaluate the robustness of NMT models to our attack, we propose enhancements to existing black-box word-replacement-based attacks by incorporating output translations of the target NMT model and the output logits of a classifier within the attack process. Extensive experiments, including a comparison with existing untargeted attacks, show that our attack is considerably more successful in altering the class of the output translation and has more effect on the translation. This new paradigm can reveal the vulnerabilities of NMT systems by focusing on the class of translation rather than the mere translation quality as studied traditionally.

\end{abstract}

\section{Introduction}
Recently, deep neural networks have emerged as powerful tools in various domains, such as  Natural Language Processing (NLP) \cite{vaswani2017attention} and computer vision \cite{he2016deep}. Despite their exceptional performance, these models have been shown to be susceptible to slight perturbations to their inputs, known as adversarial attacks \cite{szegedy2014intriguing, moosavi2016deepfool, madry2018towards}. In particular, adversarial examples closely resemble the original input and can deceive the target model to generate incorrect outputs. Extensive research has also been devoted to adversarial attacks against NLP models \citep{jin2020bert,li2020bert,zang2020word,guo2021gradient,wang2022semattack,zou2023universal} since NLP models are increasingly employed in practical systems.
These studies have mainly focused on \textit{text classification} tasks such as sentiment classification and natural language inference. In text classification, the adversary aims to fool the target model into misclassifying the input sentence as a specific wrong class (targeted attacks) or any class other than the correct ground-truth class (untargeted attacks).


\begin{table*}[t]
	\centering
		\renewcommand{\arraystretch}{0.95}

	\setlength{\tabcolsep}{2.4pt}
	
	\scalebox{0.78}{

		\begin{tabular}[t]{@{} l| >{\parfillskip=0pt}p{6.5cm}| >{\parfillskip=0pt}p{7cm}|c|c @{}}
			\toprule[1pt]
		     
		      & \multirow{1}{*}{\textbf{Input Text} (\textcolor{red}{red}=adversarial perturbation)} & \multirow{1}{*}{\textbf{En-to-Fr NMT} (\textcolor{blue}{blue}=English meaning)} & \textbf{Classifier} & \textbf{True Class} \\
		
			\midrule[1pt]
			
		     Org. &  unflinchingly bleak and desperate \hfill\mbox{} &   Inébranlablement sombre et désespéré \hfill\mbox{} & Neg. & Neg. \\
			 

			 Adv. ACT\textsubscript{TF} &  unflinchingly bleak and \textbf{\textcolor{red}{upsetting}} \hfill\mbox{} &  Légère et bouleversante, sans fin  \hfill\mbox{} & Pos. & Pos. \\
    &   &  (\textcolor{blue}{Light and moving, endless})  \hfill\mbox{} &  &  \\


   Adv. TF & unflinchingly \textbf{\textcolor{red}{melancholy}} and \textbf{\textcolor{red}{upsetting}} \hfill\mbox{} & Mélancolie et bouleversement inébranlables  \hfill\mbox{} & Pos. & Neg.  \\
   &   &  (\textcolor{blue}{Unwaveringly melancholy and turmoil})  \hfill\mbox{} &  &  \\

   \bottomrule[1pt]
			

		\end{tabular}
	}
	\caption{\fontsize{9.8}{11}\selectfont{Illustration of valid adversarial attacks with two examples against Marian NMT (En-Fr) on a movie review from SST-2 dataset. For both examples, the class predicted by the classifier for the adversarial translation differs from the class of the original translation. The first adversarial example, ACT\textsubscript{TF}, is considered valid since the NMT's French translation is wrong (if we compare its meaning in blue to the input text), and the classifier's prediction is correct. However, in the second one, TF, the NMT's translation is correct, and the classifier's prediction is wrong. Hence, it is not a desirable  attack. The last column, True Class, is the real perceived sentiment of the translated review. }}
	\label{tab:issue}
	
\end{table*}

Another important task in NLP is \textit{Neural Machine Translation (NMT)}, which has gained significant attention \cite{bahdanau2015neural}.  In this application also, adversarial attacks have been studied to gain insights into the vulnerabilities of these systems. Particularly, untargeted attacks seek to generate adversarial examples that preserve the semantics in the source language while the output translation by the target model is far from the true translation \cite{ebrahimi2018adversarial,cheng2019robust,michel2019evaluation,niu2020evaluating,zou2020reinforced,sadrizadeh2023transfool}. On the other hand, targeted attacks against NMT systems aim to mute or push specific target words in the translation \cite{ebrahimi2018adversarial,cheng2020seq2sick,wallace2020imitation,sadrizadeh2023targeted}. 
None of these attacks against NMT systems actually consider the class of the output translation as the objective of the adversarial attack. However, in some cases, the user only cares about the class, e.g., sentiment, of the translation rather than the exact translation. Moreover, the class of the translation reflects the whole meaning of the sentence.
Nevertheless, simply reducing the translation quality (untargeted attacks) or inserting specific keywords in the translation (targeted attacks), as proposed in previous works, may not sufficiently alter the translation and thus change the overall category of the translation. Moreover, in previous attack frameworks, it is particularly challenging to evaluate the true impact of the adversarial attack on the performance of the target NMT  since the ground-truth translation might change even if the adversarial perturbation is subtle \cite{zhang2021crafting,sadrizadeh2023transfool}.

In light of the above challenges, this paper introduces ACT (Altering Class of Translation), a novel adversarial attack framework against NMT systems guided by a classification objective. 
In our attack strategy, the adversary aims to craft an adversarial example in the source language that deceives the \textit{target NMT model}. Specifically, the goal is to make the  translation of the adversarial sentence belong to a different class than the original translation. To achieve this, the adversary uses an \textit{arbitrary classifier as an oracle} to  predict and alter the class of the output translation by the target NMT model. By targeting the class of the  output translation, our attack has a more substantial effect on the translation. The second row of Table \ref{tab:issue} shows a successful adversarial example generated in our framework. The original movie review has negative sentiment. The adversary aims to change the input sentence (in an imperceptible manner, i.e., the sentiment remains negative) such that the output translation by the target NMT model has positive sentiment instead of negative as the original translation. However, the second example (last row) is an undesirable case, where the class of the translation predicted by  the classifier  is changed, but the classifier is fooled by the attack, and its prediction is incorrect. In our framework, the target of the attack is the NMT model, but the attacker considers a system, including the target NMT model and the classifier operating on the output of the NMT model. Hence, the attacker needs to ensure that the attack fools the NMT model (not the classifier) to generate a wrong translation, which then results in a different class predicted by the classifier.  Consequently, the main challenge in this framework is distinguishing the attack's impact on the target NMT model from its impact on the classifier, which is not attacked by the adversary. 
To address this challenge, we  design enhancements to existing black-box word-replacement attacks, such as TextFooler \cite{jin2020bert} and BAE \cite{garg2020bae}, by integrating the output translation of the target NMT model and the output logits of a classifier into the attack process. This approach ensures that the adversarial translation is far from the original one, in addition to  altering the class to which the translation belongs. In practice, attackers typically have limited access to the target NMT model. Therefore, we assume a black-box setting for the attack against the target NMT model. Additionally, we assume black-box access to the oracle classifier, as the adversary may employ  an off-the-shelf classifier to guide the attack.


We extensively evaluate the robustness of NMT systems to our proposed attack. 
To evaluate the attack, we measure the success rate of altering the class of the output translation by using a classifier that was not involved in the attack process. Moreover, we estimate the impact of the attack on the NMT model by the similarity between the translations of  the original and adversarial sentences.  
As a baseline, we consider the untargeted attacks against NMT systems and check if the class of the translation changes after the attack. Our experiments show that, although untargeted attacks can reduce the translation quality, they are notably less successful in changing the category of the translation. In contrast, our proposed attack not only changes the class of the translation but also has more impact on the translation. It shows that our new attack can provide a more comprehensive evaluation of the robustness of  NMT systems to adversarial attacks.  In summary, our contributions are as follows:

\begin{itemize}[leftmargin=*]
\setlength\itemsep{0.01mm}
    \item  We introduce ACT, a novel attack framework against NMT models, which is guided by a classifier to change the class of the output translation. 
    \item We propose modifications to existing black-box word-replacement attacks to make them effective for the proposed   attack strategy.
    \item We thoroughly assess the   robustness  of NMT systems to our new attack framework, which showcases the  vulnerabilities of NMT systems by focusing on the class of output translation.
\end{itemize}



\section{Related Works}
Textual adversarial attacks pose unique challenges compared to their image domain counterparts. The discrete nature of textual data makes it challenging to employ gradient-based optimization methods directly. Moreover, defining the imperceptibility of adversarial perturbations for text is difficult. While a significant portion of research has focused on text classification, some works also target sequence-to-sequence NLP systems, such as NMT.
\paragraph*{Text Classification}

In the text classification tasks, some attacks measure imperceptibility based on the number of edits at the character level \cite{ebrahimi2018hotflip,gao2018black,pruthi2019combating}. However,  most NLP attacks consider semantic similarity as the imperceptibility metric and operate at the word level. Some of these attacks adopt optimization methods to bridge the gap between image and textual domains   \cite{,guo2021gradient,sadrizadeh2022block,yuan-etal-2023-bridge}. However, the majority of methods select specific words in the input sentence and replace them with synonyms \cite{zang2020word}, similar words in the embedding space \cite{alzantot2018generating,ren2019generating,jin2020bert,maheshwary2021generating,ye2022texthoaxer}, or candidates predicted by a masked language model \cite{garg2020bae,li2020bert,li2021contextualized,yoo2021towards}. 



\paragraph*{Neural Machine Translation}

In contrast to text classification settings, where the adversary aims to alter the predicted class, NMT models generate entire sentences as their output and the adversary tries to alter this translation.  There are various types of adversarial attacks in the literature depending on the adversary's objective. Untargeted attacks aim to reduce the translation quality of the target model with respect to the ground-truth translation \cite{michel2019evaluation,cheng2019robust,cheng2020advaug,sadrizadeh2023transfool}. Targeted attacks seek to mute or insert specific words into the translation \cite{ebrahimi2018adversarial,cheng2020seq2sick,wallace2020imitation,sadrizadeh2023targeted}. \cite{wallace2020imitation} introduces a universal attack that causes incorrect translation by target model when a single snippet of text is appended to any input sentence. They also show that NMT models can generate malicious translations from gibberish input. 

Specifically, \cite{belinkov2018synthetic,ebrahimi2018adversarial} first explore the vulnerabilities of NMT models to character manipulations. In untargeted attacks, \cite{cheng2019robust} replaces random words in the input sentence with the words suggested by a language model, guided by gradients to reduce translation quality. Moreover, \cite{michel2019evaluation,zhang2021crafting} substitute important words in the input sentence with their neighbouring words in the embedding space.  Other approaches utilize optimization to generate adversarial examples \cite{cheng2020seq2sick,sadrizadeh2023transfool}. While the first use the NMT embeddings to define similarity, the latter uses the embedding representation of a language model. 


However, none of these attacks specifically target the class that the translation belongs to, which can be important in many applications. Furthermore, evaluating the true impact of these attacks is challenging, as the adversarial perturbations may change the ground-truth output and potentially result in an overestimation of the attack performance. To address these limitations, we propose a novel attack framework against NMT models guided by a classifier to generate an adversarial example whose translation by the NMT model belong to a different class than the class of the original translation. This approach can have more impact on the output translation by altering its class. We should note that, in a parallel work, \cite{raina2023sentiment} recently published a paper proposing an attack against NMT models to change the perception of translation. In contrast to this work, we consider modifying the class of output translation, and not just sentiment. Our proposed framework can be used with different classifiers based on the adversary's objective. Moreover, we try to distinguish the attack's impact on the target NMT model from its impact on the classifier used in the attack, which is the main challenge in this framework. Finally, we extensively evaluate the robustness of NMT models to our attack framework. We discuss these differences in more detail in Appendix \ref{discussion}. 


\section{ACT Attack Framework against NMT}
In this section, we introduce ACT, our new  attack framework against NMT models. Then, we present our method to craft such adversarial examples.

\subsection{Attack Definition} \label{attack_def}

\begin{figure}[t]
\centering
\includegraphics[width=1\columnwidth]{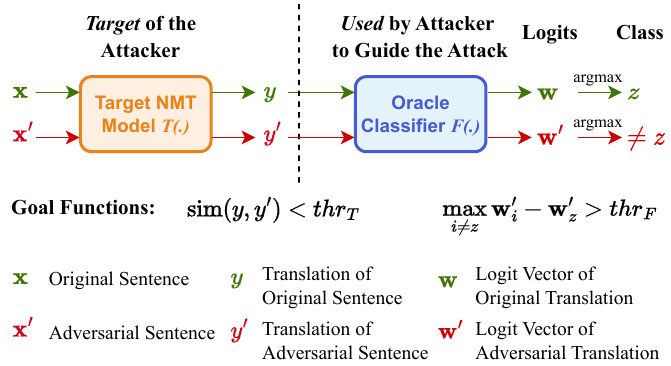} 
\caption{Block diagram of ACT, a new attack framework against NMT models.}
\label{fig:block}
\end{figure}

The block diagram of our attack framework is presented in Figure \ref{fig:block}.  Let $\mathcal{X}$  denote the source language space and $\mathcal{Y}$  denote the target language space.
The NMT model $T: \mathcal{X}\rightarrow \mathcal{Y}$ maximizes the probability of generating the true translation,  automatically translating  the input sentence $\mathbf{y} = T(\mathbf{x})$, where $\mathbf{x} \in \mathcal{X}$ is the input sentence in the source language, and $\mathbf{y} \in \mathcal{Y}$ is the output translation by the NMT model $T$. 

In our proposed attack, the adversary seeks an adversarial example $\mathbf{x'}$ in the source language that misleads the \textit{target NMT model}. In particular, the adversary aims that the adversarial translation $\mathbf{y'} = T(\mathbf{x'})$ is from a different class than the original translation. To this end, the adversary uses an \textit{arbitrary classifier as an oracle} $F: \mathcal{Y}\rightarrow \mathcal{Z}$ to determine the class of the output translation by the target NMT model.\footnote{The effect of the choice of classifier on the performance of ACT is studied in Appendix \ref{class}.} Based on the attack objective, the adversary can use a classifier suitable for any task, such as sentiment classification. The classifier classifies the input translation into $z = F(\mathbf{y})$, where $z$ is the class of the translation $\mathbf{y}$. Hence, the adversarial example is crafted such that $F(\mathbf{y'}) \neq z$. However, we constraint that the  adversarial example $\mathbf{x'}$ must remain a   natural  sentence and be semantically similar to the original sentence $\mathbf{x}$.

In practice, the adversary has limited access to the target model. Therefore, we consider a black-box scenario where the adversary cannot access the model parameters, architecture,  or training data of the target NMT model $T$. Moreover, the adversary may use an off-the-shelf classifier for the attack and hence, has black-box access to the classifier $F$. The attacker can only query the NMT model $T$ with a sentence in the source language and get the translated sentence. Then, they can use the classifier $F$ to determine the class and the corresponding logits for the translated sentence. 

It is worth mentioning that we can consider other scenarios within our framework, e.g., when the target of the attack is the entire system (NMT + classifier), or just the classifier. We explore these scenarios in the experiments. 

\subsection{Methodology}
There are substantial textual adversarial attacks in the literature that efficiently search the discrete space of tokens to craft meaning-preserving natural adversarial sentences \cite{jin2020bert,garg2020bae,ye2022texthoaxer}. We can build our attack upon these attacks based on the objective of our attack framework. To this end, we use TextAttack, which provides a unified framework for numerous textual adversarial attacks \cite{morris2020textattack} and facilitates the incorporation of our enhancements into the existing attack methodologies. In this framework, there are four components in black-box attacks based on word-replacement. 

\paragraph*{Constraints:} In order to generate semantic-preserving and grammatically correct adversarial sentences, each attack defines a set of constraints by using a grammar checker, embedding space distance, or perplexity score.

\paragraph*{Transformation:} To find a set of candidates to replace the selected words, various transformations are proposed, e.g., predictions  by masked language models or neighbors in the embedding space.

\paragraph*{Search method:} Each attack employs a search method to iteratively query the target model and find an adversarial example that satisfies the constraints, e.g., greedy search or genetic algorithm. 

\paragraph*{Goal function:} This module specifies the stopping criteria  and determines if an attack is successful.

In the attacks modeled by this framework, we can craft an adversarial example that satisfies the adversary's goal function and adheres to certain linguistic constraints. We find these perturbations by replacing some of the words in the input sentence based on a search method. 

In our proposed attack framework, the adversarial sentence should maintain the semantics and the class label in the source language. Meanwhile, the translation in the target language by the target NMT model should have a different class label. Therefore, we use the \textit{constraints} and \textit{transformations} in the literature to generate adversarial examples that are similar to the original sentence and are grammatically correct. However, based on our attack objective, we propose a new goal function and some alterations to the search method. In our framework, the adversary uses an oracle classifier to manipulate the NMT model into crafting an adversarial translation from a different class. Hence, the target of the attack is the NMT model, and we need to distinguish  between the impact of the attack on the NMT model $T$ and on the classifier $F$. An undesirable example, in which the classifier is impacted by the attack, is presented in Table \ref{tab:issue}. To address this challenge and make the attack mainly effective on the NMT model rather than the classifier, we propose  the two following goal functions as the stopping criteria of the attack:
\paragraph*{1) Translation}
In order to highlight the effect of the attack on the NMT model, we include the translation output of the NMT model in the goal function. In other words, we consider an adversarial attack to be successful if the similarity between the original translation $\mathbf{y}$ and the adversarial translation $\mathbf{y'}$ is less than a threshold $thr_T$:

\begin{equation}\label{trQ}
    \text{sim}(\mathbf{y}, \mathbf{y'})< thr_T.
\end{equation}
This goal function allows us to mislead the NMT model $T$ to generate a translation for the adversarial example that is far from the original translation. We use the BLEU score to evaluate the similarity between two
translations since it is fast, common in benchmarks, and has been used in previous works \cite{cheng2019robust,wallace2020imitation,zhang2021crafting}.\footnote{We use case-sensitive SacreBLEU on detokenized sentences.} We study the effect of this similarity metric when we use BLEURT-20 \cite{sellam2020learning} instead of BLEU score in Appendix \ref{sim}.

\paragraph*{2) Classification} \looseness=-1 Instead of checking whether the classifier's output is different from the ground-truth label, we use the raw output of the classifier before the softmax, known as \textit{logit}. In order to ensure that the output of the classifier is different from the original class with high confidence,  we consider an adversarial attack to be successful if the difference between the logits of the most
probable class and the ground-truth class is larger than a threshold $thr_F$:
\begin{equation}\label{logit}
    \max\limits_{i\neq z}{\mathbf{w}'_i} - \mathbf{w}'_z > thr_F,
\end{equation}
where $\mathbf{w'} = W(\mathbf{y'})$ are the logits, and $z$ is the ground-truth class. 
Therefore, we consider an adversarial example to be successful if both goal functions \eqref{trQ} and \eqref{logit} are satisfied.

On another note, most of the existing attacks use a score function during the search to estimate the importance of  the tokens in the sentence. They select the important words in the sentence to  limit the search space and the number of alterations made by the attack \cite{ren2019generating,li2020bert,jin2020bert,garg2020bae}.  
They define this score function as the  logit of the ground-truth class, i.e.,  $\mathbf{w}'_z$. The token importance is the decrease in this score when removing a token from the sentence. We propose a new score function to account for  the effect of the adversarial example on the NMT model as follows:

\begin{equation}\label{score}
    S(\mathbf{x'}) = \mathbf{w}'_z + \alpha \text{ sim}(\mathbf{y},\mathbf{y'}). 
\end{equation}
The proposed second term in this score function makes the importance of the token dependent on the decision of the \textit{classifier} \textit{and} that of the \textit{target NMT model}.

\begin{table*}[t]
	\centering
		\renewcommand{\arraystretch}{0.9}
	\setlength{\tabcolsep}{4pt}
	
	\scalebox{0.88}{
		\begin{tabular}[t]{@{} ccccccccccccccc @{}}
			\toprule[1pt]
		    \multirow{2}{*}{\textbf{Task}}  &
		    \multirow{2}{*}{\textbf{Method}}  & \multicolumn{6}{c}{\textbf{Marian NMT (En-Fr)}} &&   \multicolumn{6}{c}{\textbf{Marian NMT (En-De)}} \\
			\cline{3-8}
			\cline{10-15}
			\rule{0pt}{2.5ex}    
			& & \scalebox{0.95}{ASR$\uparrow$} & \scalebox{0.95}{BLEU$\downarrow$} & \scalebox{0.95}{chrF$\downarrow$} & \scalebox{0.95}{Sim.$\uparrow$} & \scalebox{0.95}{Perp.$\downarrow$} & \scalebox{0.95}{WER$\downarrow$} 
            && \scalebox{0.95}{ASR$\uparrow$} & \scalebox{0.95}{BLEU$\downarrow$} & \scalebox{0.95}{chrF$\downarrow$} & \scalebox{0.95}{Sim.$\uparrow$} & \scalebox{0.95}{Perp.$\downarrow$} & \scalebox{0.95}{WER$\downarrow$}\\ 
			\midrule[1pt]
		\multirow{5}{*}{\rotatebox[origin=c]{90}{SST-2}} 
		& kNN &   8.95 & 50.95 & 70.53 & 0.78 & 6.06 & 30.46 
  && 
            12.08 & 42.81 & 69.18 & 0.82 &  5.19 & 28.02 
            \\
		& Seq2Sick & 2.14 & 41.83 & 59.52 & 0.60 & 4.54 & 35.56 
  &&  2.71 & 46.39 & 60.01 & 0.71 &  3.54 & 22.48 
  \\
            & TransFool &  10.21 & 44.90 & 68.16 & \textbf{0.85} &  2.64 & 20.86 
            && 12.35 & 39.12 & 62.85 & 0.82 & 3.04 & 21.82 
            \\ 
            & ACT\textsubscript{TF} & \textbf{40.23} & \textbf{29.66} & \textbf{56.11} & 0.84 & 2.00 & 21.19 
            && \textbf{40.84} & \textbf{25.11} & \textbf{53.84} & 0.84 &  2.23 & 22.13 
            \\
            & ACT\textsubscript{BAE} & 38.08 & 45.72 & 65.62 & \textbf{0.85} & \textbf{0.59} & \textbf{15.16} 
            && 31.61 & 40.95 & 64.80 & \textbf{0.86} &  \textbf{0.61 } & \textbf{14.95} 
            \\
		\midrule[1pt]
		\multirow{5}{*}{\rotatebox[origin=c]{90}{MR}} 
		  & kNN &  8.78 & 50.01 & 70.93 & 0.82 & 6.88 & 31.35 
    && 13.00 & 44.22 & 70.77 & 0.83 &  4.86 & 27.10 
    \\
		& Seq2Sick &  3.80 & 30.82 & 53.44 & 0.71 & 5.86 & 38.78 
  &&  3.00 & 26.40 & 56.50 & 0.73 &  2.48 & 21.25 
  \\
            & TransFool & 10.56 & 43.25 & 66.16 & 0.84 & 3.04 & 24.81 
            && 14.63 & 37.47 & 66.53 & 0.85 & 2.76 & 22.54 
            \\ 
            & ACT\textsubscript{TF} & \textbf{36.37} & \textbf{20.15} & \textbf{48.46 }& 0.82 & 3.03 & 26.20 
            && \textbf{26.50} & \textbf{9.63} & \textbf{42.62} & 0.80 &  3.71  & 29.67 
            \\
            & ACT\textsubscript{BAE} & 31.64 & 46.08 & 65.26 & \textbf{0.86} & \textbf{0.62} & \textbf{15.19} 
            && 21.00 & 35.51 & 62.68 & \textbf{0.88} &  \textbf{0.63} & \textbf{14.74} 
            \\
		\midrule[1pt]
		\multirow{4}{*}{\rotatebox[origin=c]{90}{\scalebox{0.95}{AG's News}}} 
		& kNN &   2.02 & 40.81 & 69.50 & \textbf{0.95} & 
            2.13 & \textbf{10.67} 
            && 2.65 & 65.49 & 81.70 & \textbf{0.95} & 1.58 & \textbf{9.67} 
            \\
            & TransFool &  3.30 & 48.61 & 68.09 & 0.89 & 3.07 & 17.89 
            && 3.43 & 44.74 & 66.27 & 0.90 & 2.65 & 18.35 
            \\ 
            & ACT\textsubscript{TF} & \textbf{22.84} & \textbf{25.62} & \textbf{49.19} & 0.85 & 6.02 & 27.50 
            && \textbf{18.36} & \textbf{27.03} & \textbf{52.08} & 0.85 &  5.87  & 27.28 
            \\
            & ACT\textsubscript{BAE} & 7.58 & 36.10 & 62.11 & 0.94 & \textbf{1.31} & 12.91 
            && 7.19 & 42.24 & 66.59 & 0.93 & \textbf{1.49} & 14.37 
            \\
			
			\bottomrule[1pt]
		\end{tabular}
  }
  \caption{Evaluation results of the adversarial attacks against two translation tasks and three different datasets.  
	} \label{tab:marian}
	
\end{table*}

\section {Experiments}

In this section, we discuss our experimental setup, and then we conduct comprehensive experiments to evaluate the robustness of  various NMT models and tasks in the face of our proposed attack strategy.\footnote{Our source code is available at \url{https://github.com/sssadrizadeh/ACT}.}

\subsection{Experimental Setup}
We evaluate the robustness of transformer-based NMT models to our attack. Specifically, we target the HuggingFace implementation of Marian NMT models \cite{junczys-dowmunt-etal-2018-marian} and mBART50 multilingual NMT model \cite{tang2020multilingual} to validate the effectiveness of our attack across diverse architectures. Moreover, we conduct experiments on the English-French (En-Fr) and English-German (En-De) translation tasks.

In our proposed attack strategy, the adversary aims to alter the class of the NMT model's output translation. To achieve a comprehensive evaluation, we require ground-truth class information for the sentences. Therefore, instead of translation datasets, we consider text classification datasets, including SST-2 \cite{socher2013recursive}, MR \cite{pang2005seeing}, and AG’s News \cite{zhang2015character}. SST-2 and MR are sentiment classification datasets, while AG’s News is a topic classification dataset. We perform the attack on the test set of these datasets.\footnote{For the AG's News and MR datasets, we attack the first 1000 sentences from the test set.}

We translate the training set of these datasets using the target NMT model and fine-tune two separate classifiers.\footnote{For the En-Fr task, we fine-tuned two models: \url{https://huggingface.co/asi/gpt-fr-cased-small} with GPT-2 architecture and \url{https://huggingface.co/tblard/tf-allocine} with BERT architecture. As for the En-De task, we fine-tuned two additional models: \url{https://huggingface.co/dbmdz/german-gpt2} with GPT-2 architecture and \url{https://huggingface.co/oliverguhr/german-sentiment-bert} with BERT architecture.}  We utilize one of the classifiers during the attack process, while the other one is reserved for evaluation. This approach ensures fair comparisons with the baselines and accounts for the possibility of the adversarial attack fooling the classifier used in the attack. More details about the datasets and models are reported in Appendix \ref{model}.

To perform the proposed attack, we use TextAttack implementation of TextFooler (TF) \cite{jin2020bert} and BAE \cite{garg2020bae}. We change the goal function and the score function, as explained in the last section. As for the parameters, we set $thr_T=0.4$, $thr_F=2$ and $\alpha=3$ based on the ablation study available later.

As a baseline, we compare with the untargeted attacks against NMT systems and examine whether the class of the translation changes after the attack. Specifically, we compare with the \textit{kNN} attack from \cite{michel2019evaluation}, which is a white-box untargeted attack against NMT
models that substitutes some words with their neighbors in the embedding space. Additionally, we adapt the targeted attack \textit{Seq2Sick} \cite{cheng2020seq2sick}, which is based on  optimization in the NMT embedding space, to the untargeted setting. Lastly, we compare with \textit{TransFool},  an untargeted attack against NMT models that is also based on optimization but uses language model embeddings to preserve the semantics.

For evaluation, we report several performance metrics. We measure  \textit{Attack Success Rate (ASR)} of the adversarial examples by testing them on a classifier that was not involved in the attack process.  We also measure the similarity between the translations of the original and adversarial sentences using  BLEU score and chrF \cite{popovic2015chrf}.  A lower similarity  indicates a greater deviation between the translations of the adversarial  and  original sentences, which allows us to estimate  the impact of the attack  on the target NMT model. Furthermore, we use Universal Sentence Encoder (USE) \cite{yang2020multilingual} to approximate the semantic Similarity (Sim.) between the original and adversarial sentences. To measure the naturality of the adversarial sentences, we calculate the relative increase in the  Perplexity score (Perp.) of GPT-2 (large) between the adversarial and original sentences. Finally, we report Word Error Rate (WER), i.e., the percentage of words that are modified by an adversarial attack. 

\begin{table*}[t]
	\centering
		\renewcommand{\arraystretch}{1.}

	\setlength{\tabcolsep}{2pt}
	
	\scalebox{0.87}{

		\begin{tabular}[t]{@{} c|l| >{\parfillskip=0pt}p{13.5cm} @{}}
			\toprule[1pt]
		    \textbf{Task} & 
		    \textbf{Sentence}  & 
      \textbf{Text}\\
		
			\midrule[1pt]
			
		    \multirow{5}{*}{\rotatebox[origin=c]{90}{MR}}  & Org. &  
      a solidly entertaining little film . \hfill\mbox{}  \\
			 

			& Org. Trans. \hfill (Positive) &  
   un petit film très divertissant. \hfill\mbox{}  \\
			 
			\cline{2-3}
			\rule{0pt}{2.5ex}

			& Adv. ACT\textsubscript{TF} &  
   a solidly \textbf{\textcolor{red}{goofy}} little film . \hfill\mbox{} \\
			
			
		&	\multirow{1}{*}{Adv. Trans. \hfill (Negative)} &  
   un petit film complètement dégueulasse. {\textcolor{blue}{(\textit{dégueulasse means 'nasty', which is  negative.})}} \hfill\mbox{} \\

			\cline{1-3}
			\rule{0pt}{2.5ex} 
			
	\multirow{8}{*}{\rotatebox[origin=c]{90}{SST-2}} &	    \multirow{2}{*}{Org.} &  
      the primitive force of this film seems to bubble up from the vast collective memory of the combatants . \hfill\mbox{}  \\
			 

		&	\multirow{1}{*}{Org. Trans. \hfill (Positive)} &  
 la force primitive de ce film semble jaillir de la vaste mémoire collective des combattants. \hfill\mbox{}  \\
			 
			\cline{2-3}
			\rule{0pt}{2.5ex}

		&	\multirow{2}{*}{Adv. ACT\textsubscript{BAE}} &  
     the primitive \textbf{\textcolor{red}{tone}} of this film seems to bubble up from the vast collective \textbf{\textcolor{red}{memories}} of the combatants .   \hfill\mbox{} \\
			
			
		&	\multirow{2}{*}{Adv. Trans. \hfill (Negative)} &  
   Le ton primitif de ce film semble s'estomper des vastes souvenirs collectifs des combattants. {\textcolor{blue}{(\textit{s'estomper means fade which conveys negative meaning})}} \hfill\mbox{} \\

			\bottomrule[1pt]
			

		\end{tabular}
	}
	\caption{Adversarial examples against Marian NMT (En-Fr).}
	\label{tab:sample}
	
\end{table*}

\subsection{Results}

Now we evaluate the robustness of various NMT models to our new attack strategy. Table \ref{tab:marian} shows the performance of different attacks against Marian NMT models for (En-Fr) and (En-De) tasks. Additionally, the performance against mBART50 NMT model for (En-Fr) task is reported in Appendix \ref{mbart}. In these tables, ACT\textsubscript{TF} and ACT\textsubscript{BAE} denote the modified version of the corresponding attacks with our proposed changes. These results demonstrate that the existing untargeted adversarial attacks against NMT models can generate adversarial examples with translations dissimilar to the original translation. However, they have a low success rate in changing the class of the translation. Seq2Sick has the lowest success rate, while TransFool is the most successful one. On the other hand, ACT\textsubscript{TF} and ACT\textsubscript{BAE} have much higher success rates than the baselines in all cases, and they are able to change the class of the NMT models' output translations. Interestingly, these two methods, especially ACT\textsubscript{TF}, can generate adversarial examples whose translations are further away from the original translation than those generated by the baselines, i.e., lower BLEU score and chrF. While ACT\textsubscript{TF} has a higher success rate and causes more damage to the translation, it generates adversarial examples with higher perplexity scores and lower similarity than the ones generated by ACT\textsubscript{BAE}. It is worth noting that all attacks are much less successful in  AG's News. It seems that when the number of classes is larger, the attack becomes more challenging. 
We should note that Seq2Sick is not successful against AG's News and  not reported in  Tables \ref{tab:marian}.

Regarding the run-time, for the Marian NMT (En-Fr) model and SST-2 dataset, on a system equipped with two NVIDIA A100 GPUs, it takes 26.54 and 15.75 seconds to generate adversarial examples by ACT\textsubscript{TF} and ACT\textsubscript{BAE}, respectively. If we do not use the proposed modifications, the run-time of ACT\textsubscript{TF} would be 17.17 seconds. Table \ref{tab:sample} shows some adversarial examples generated by ACT\textsubscript{TF} and ACT\textsubscript{BAE}. These samples show that while the proposed attack maintains the semantic similarity in the source language, they are able to force the NMT model to generate a translation from a different class in the target language. More adversarial examples can be found in Appendix \ref{example}.

All in all, previous untargeted adversarial attacks are not much successful in deceiving the NMT model to generate translations from a different class than the original translations. However, this type of attack can be   more harmful to the users since the overall meaning of the translation is changed. The proposed attacks, i.e., ACT\textsubscript{TF} and ACT\textsubscript{BAE}, are more successful in changing the class of the adversarial translation. Moreover, compared to baselines, the adversarial translations are further away from the original translation.

We should note that, in our framework, we are  generating adversarial examples that are robust to the translation. It has been shown in \cite{bhandari2023lost} that most of the adversarial attacks against text classifiers are not robust to translation. This means that most of the attacks in the source language do not transfer to the translation model. Therefore, even if the attacker changes the class in the source language, it is possible that the output translation is still from the correct class. 

\begin{table}[!t]
	\centering
		\renewcommand{\arraystretch}{1}
	\setlength{\tabcolsep}{2.4pt}

\scalebox{0.9}{
		\begin{tabular}[t]{@{} ccccccccc @{}}
			\toprule[1pt]
		    \multirow{2}{*}{\textbf{Task}}  &
		    \multicolumn{2}{c}{\textbf{kNN}} &&   \multicolumn{2}{c}{\textbf{Seq2Sick}} && \multicolumn{2}{c}{\textbf{TransFool}} \\
      \cline{2-3}
      \cline{5-6}
      \cline{8-9}
		    & BLEU$\downarrow$ & chrF$\downarrow$ && BLEU$\downarrow$ & chrF$\downarrow$ && BLEU$\downarrow$ & chrF$\downarrow$\\
			\midrule[1pt]
			\multirow{2}{*}{\rotatebox[origin=c]{90}{SST-2}} &  47.86 & 70.68 && 93.39 & 95.95 && 43.73 & 65.79\\ 
                &  34.57 & 60.14 && 61.20 & 70.93 && 37.18 & 58.00 \\ 
            \midrule[1pt]
			\multirow{2}{*}{\rotatebox[origin=c]{90}{MR}} &   45.69 & 69.11 && 93.72 & 95.93 && 43.20 & 65.67 \\  
                &  32.13 &  57.18 && 43.48 & 62.01 && 39.17 & 61.92 \\

			\bottomrule[1pt]
		\end{tabular}
  }
  \caption{\fontsize{9.4}{11}\selectfont{Translation performance of baseline  attacks against mBART50 (En-Fr). First rows show the results for all of the adversarial examples, while the second rows correspond to the successful examples that \textit{change the class}.}} \label{tab:trans}
\end{table} 

\subsection{Analysis}
\looseness=-1 In this section, we analyze the significance of the proposed goal functions and discuss two other scenarios of the classification-guided attack strategy. We also study the transferability of the proposed attack to other NMT systems, the extension to the targeted settings, and the effect of different parameters on our attack  in Appendices \ref{transfer}, \ref{target} and \ref{param}, respectively. 
Finally, in Appendix \ref{ensemble} we show that by using an ensemble of classifiers, we can improve the performance.

\paragraph*{Impact of Classification-Guided Strategy on Translation} 
First, we show that by changing the class of the output translation, the adversary can have more impact on the NMT model. Table \ref{tab:trans} reports the similarity between the original and adversarial translation, as an estimate of the effect of the adversarial attack on the NMT model, for the baselines when for all the adversarial examples (first row) compared to the successful ones that change the class of the output translation. Across various methods and tasks, we can see that when adversarial examples change the translation's class, their translations are less similar to the original translations. This difference in similarity arises because these examples often change the overall meaning of the translation.

\begin{table}[t]
	\centering
		\renewcommand{\arraystretch}{0.9}
	\setlength{\tabcolsep}{2.3pt}
 \scalebox{0.97}{
		\begin{tabular}[t]{@{} lccccc @{}}
			\toprule[1pt]
		    \multirow{1}{*}{\textbf{Goal Func.}}  &
		    ASR$\uparrow$ & BLEU$\downarrow$ & chrF$\downarrow$ & Sim.$\uparrow$ & Perp.$\downarrow$ \\
			\midrule[1pt]
			\multirow{1}{*}{Label change} &  28.12 & 58.16 & 72.90 & 0.89 & 0.91  \\ 
			\multirow{1}{*}{Logit dif. } &   37.70 & 52.19 & 68.35 & 0.87 & 1.26 \\  
                \multirow{1}{*}{Trans. sim. } &   36.19 & 31.90 & 58.24 & 0.85 & 1.83 \\ 
                \multirow{1}{*}{ACT\textsubscript{TF}} &   40.23 & 29.66 & 56.11 & 0.84 & 2.00  \\

			\bottomrule[1pt]
		\end{tabular}
  }
  \caption{\fontsize{9.7}{11}\selectfont{Ablation study on the proposed goal functions for ACT\textsubscript{TF}  against Marian NMT (En-Fr)  on SST-2 dataset.}} \label{tab:abl}





\end{table} 

\paragraph*{Goal Functions}
Our framework consists of a classifier that acts on the output translation by the target NMT model. To ensure that the attack's influence on the target NMT model outweighs its effect on the classifier, we proposed two goal functions based on the output translation of the NMT model and the output logits of the classifier. Table \ref{tab:abl} shows the effect of these two goal functions on the attack performance. The first row shows the results when the goal function of the attack is only to change the label of the translation. The second and third rows  present the effect of using equations \eqref{logit} and \eqref{trQ}, respectively. We can see that both of the proposed goal functions increase the success rate, and they also  help to reduce the similarity between the original and adversarial translations. Since the goal function becomes more difficult to achieve, there is a decrease in  semantic similarity and an increase in the perplexity score.


\paragraph*{Other Scenarios}
So far, we have assumed that the attacker's objective is to mislead the NMT to generate a translation from a class different from the original translation. However, our proposed classification-guided strategy can be adapted to other scenarios as well. 

First, we can consider a system including a classifier that operates on the output of an NMT model.\footnote{An example might be when we are interested in the class prediction of foreign language sentences, but a classifier is available in another language. Hence, we  use an NMT model to translate the sentences before feeding them to the classifier.} In this context, the goal would be to attack the \textit{entire system} instead of just the NMT model.  This scenario is much easier than the previous one since the adversary can access the entire system. Also, the adversary's target is the performance of the entire system (NMT \textit{and} the classifier), unlike the original scenario, where the target is the NMT model. The performance of the attack in this scenario is presented in the first row of Table \ref{tab:scenario}. As expected, the success rate is much higher than that of the previous scenario. Moreover, the adversarial and original translations are more similar meaning that the NMT model is less affected by the attack.

\begin{table}[t]
	\centering
		\renewcommand{\arraystretch}{0.9}
	\setlength{\tabcolsep}{2.3pt}
	
\scalebox{0.92}{
		\begin{tabular}[t]{@{} lccccc @{}}
			\toprule[1pt]
		    \multirow{1}{*}{\textbf{Target}}  &
		    ASR$\uparrow$ & BLEU & chrF & Sim.$\uparrow$ & Perp.$\downarrow$ \\
			\midrule[1pt]
			\multirow{1}{*}{NMT + Classifier} & 95.61 & 59.08 & 73.61 & 0.89 & 0.98  \\ 
			\multirow{1}{*}{Classifier} &   57.55 & 71.92 & 83.03 & 0.91 & 0.72 \\

			\bottomrule[1pt]
		\end{tabular}
  }
  \caption{Performance of ACT\textsubscript{TF} against Marian NMT (En-Fr) on SST-2 dataset when the target is the entire system (NMT + Classifier) or just the classifier.}
	\label{tab:scenario}

\end{table} 

Secondly, we can assume that the adversary's goal is to fool only the classifier.\footnote{ This scenario may not have a practical use case, but it shows another aspect of our classification-guided attack.} Therefore, we need to craft an adversarial example whose translation is similar to the original one, the complete opposite goal of our original scenario, but the classifier predicts a wrong class for the adversarial translation. In this scenario, we can change goal function \eqref{trQ} such that the similarity is more than a threshold. Moreover, we can use equation \eqref{score} with negative coefficient $\alpha$ so that the word with a higher impact on the translation has less importance. The performance with the parameters $thr_T=0.8$ and $\alpha=-7$ are reported in the second row of Table \ref{tab:scenario}. We can see that the similarity between original and adversarial translations is  higher than those of the previous scenarios, showing that the attack is effectively targeting the classifier.

\section{Human Evaluation} \label{human}
We conduct a human evaluation campaign for the successful adversarial examples generated by ACT\textsubscript{TF} against Marian NMT (En-Fr). We randomly choose 80 successful adversarial examples on the SST-2 dataset. We split these sentences into two surveys and recruit three volunteer annotators for each survey. Since the adversarial examples are in French, we ensure that the annotators are native (mother tongue) or highly proficient in French. 

Since the naturalness and sentiment accuracy of TextFooler and BAE are already evaluated by human in their respective papers, we do not consider these two aspects of the adversarial examples in our evaluation. Instead, we study the sentiment of the adversarial translations in the target language (French). By showing the adversarial translations, we ask the annotators to choose a sentiment label from “Positive” and “Negative”. We take the majority class as the predicted label for each sentence. 

In our attack framework, the target of the attack is the NMT model, and the adversary uses a classifier to change the class of the translation. Therefore, we want the proposed attack mainly affect the NMT model rather than the classifier. Accordingly, in this study, we evaluate how much the classifier is influenced by the attack. The overall agreement between the ground-truth labels (in the dataset) and the labels predicted by the annotators is $71.3\%$. While it's true that not all adversarial translations are accurately classified by the classifier based on the annotators' labels, the majority of adversarial translations have the same sentiment as predicted by the classifier. This implies that the attack is mainly targeting the NMT model rather than fooling the classifier. Moreover, we calculate the similarity between original and adversarial translations in terms of chrF for the sentences selected for human evaluation. For sentences that the classifier's predictions align with the annotators' labels, the translation similarity is 49.46.  In contrast, for sentences that the classifier's predictions diverge from the annotators' labels, the translation similarity is 52.05. This difference highlights  that the attack is mainly affecting the NMT model rather than the classifier for the adversarial sentences whose translation deviates more from the original translation.

\section{Conclusion}
In this paper, we presented ACT, a novel adversarial attack framework against NMT models that is guided by a classifier. In our framework, the adversary aims to alter the class of the output translation in the target language while preserving semantics in the source language. By targeting the class of the output translation, we outlined a new aspect of vulnerabilities of NMT models. We proposed enhancements to existing black-box word-replacement-based attacks to evaluate the robustness of NMT models to our attack strategy. Extensive experiments and comparisons with existing untargeted attacks against NMT models showed that our attack is highly successful in changing the class of the adversarial translation. It also has more impact on the similarity of the original and adversarial translations, which highlights the potential impact of our attack strategy on the overall meaning of the NMT output translations. 

\section{Limitations}

In our framework, the target of the attack is the NMT model, and the attacker uses a classifier to change the class of the translation. Therefore, the adversary is defining  a system including the classifier operating on the output of the NMT model. Although we have proposed a goal function to make the attack mainly effective on the NMT model rather than the classifier, there is still a chance that the classifier is affected by the attack (instead of the NMT model). To consider this challenge in our evaluations, we have reported the similarity between the adversarial and original translations to measure the effect of the attack on the NMT model. Moreover, we have calculated the success rate of altering the class of translation by using a different classifier than the one used in the attack process. Such an  evaluation provides fair comparisons with the baselines and accounts for the possibility of the adversarial attack fooling the classifier used in the attack. On another note, our proposed goal functions need the output logits of the classifier. However, some of the recent textual adversarial attacks only need the hard labels \cite{ye2022texthoaxer,yu2022texthacker}. Modifying these works based on our attack framework can be explored in the future. Finally, we have considered sentiment and news classification in our experiments. It is worth considering other classification tasks, e.g., hate speech, in our framework and evaluate the robustness of NMT models in other areas.

\section{Ethic statement}
We introduced ACT, a new attack framework against NMT models, to study the vulnerabilities of NMT models from another aspect than tradition frameworks with the hope to  pave the way for building robust NMT models. Although there is a potential for malicious actors to misuse our attack, we want to emphasize that we strongly discourage the use of our method for targeting real-life NMT systems with harmful intent.
\section*{Acknowledgements}
This work has been partially supported by armasuisse Science and Technology project MULAN. 

\bibliography{main}

\clearpage
\appendix
\section*{Appendix}
\label{appendix}

In this Appendix, we first provide more details about the models and datasets used in the experiments. Afterwards, we present more experimental results of the attack with BLEURT as the similarity metric for translation, studying the choice of the classifier on the performance,  the transferability analysis of the attack, extension to the targeted settings, and more samples of the crafted adversarial examples. 
Finally, we provide a comparison with the recent parallel work of \cite{raina2023sentiment} and discuss the potential limitations of our work.

\section{Models and Datasets} \label{model}
In this Section, we provide  information about the datasets and models used in our experiments. It is worth noting that we used HuggingFace datasets \cite{wolf-etal-2020-transformers} and transformers \cite{lhoest-etal-2021-datasets} libraries.



			
			


\subsection{Target NMT Models}  
We evaluate the robustness of the HuggingFace implementation of Marian NMT models \cite{junczys-dowmunt-etal-2018-marian} and mBART50 multilingual NMT model \cite{tang2020multilingual} for En-Fr and En-De translation tasks. As a benchmark of the performance of these models, we report their translation quality on  WMT14  \cite{bojar2014findings} in table \ref{tab:dataset}.

\subsection{Datasets}

Since we require ground-truth class information for the sentences in our evaluation, instead of translation datasets, we consider text classification datasets. We use  SST-2 \cite{socher2013recursive}, MR \cite{pang2005seeing}, and AG’s News \cite{zhang2015character} in our experiments. SST-2 and MR are  Sentence-level sentiment classification datasets on positive
and negative movie reviews. On the other hand, AG’s News is a Sentence-level topic classification dataset with regard to four news topics: World, Sports, Business, and Science/Technology. Some statistics of these datasets are reported in Table \ref{tab:data}. We use the training set of these datasets to train the classifiers used for the attack and evaluation. Moreover, we use the test set to evaluate the robustness of the target NMT models, except for the SST-2, for which we used the validation set.

\subsection{Classifiers} 
We translate the training set of these datasets using the target NMT model and fine-tune two separate classifiers, employing GPT-2 \cite{radford2019language} and BERT \cite{kenton2019bert}. The accuracy of these models is reported in Table \ref{tab:data}. 

\section{Additional Results}
In this Section, we present more results of the proposed attack.

\subsection{Attack Performance against mBART50} \label{mbart}
To validate the effectiveness of our attack across diverse architectures, we also attack mBART50 NMT model. The attack performance is presented in Table \ref{tab:mbart}. These results show the same trend as that of the attack against Marian NMT model, which proes the effectiveness of our attack framework against different NMT models.
\begin{table}[t]
	\centering
		\renewcommand{\arraystretch}{1}
	\setlength{\tabcolsep}{4pt}
		\begin{tabular}[t]{@{} lccccc @{}}
			\toprule[1pt]
		    \multirow{2}{*}{\textbf{Dataset}}  &    
		    \multicolumn{2}{c}{\textbf{Marian NMT}} &&
		    \multicolumn{2}{c}{\textbf{mBART50}} \\
		    
		    \cline{2-3}
			\cline{5-6}
			\rule{0pt}{2.5ex}    
			&  BLEU & chrF && BLEU & chrF  \\
			
			\midrule[1pt]
			En-Fr  &  \multirow{2}{*}{39.88} & \multirow{2}{*}{64.94} & & \multirow{2}{*}{36.17} & \multirow{2}{*}{62.66}\\
			WMT14 \\
			\midrule
			En-De  &  \multirow{2}{*}{27.72} & \multirow{2}{*}{58.50} && \multirow{2}{*}{25.66} & \multirow{2}{*}{57.02} \\ 
			WMT14 \\
			
			\bottomrule[1pt]
		\end{tabular}
	\caption{Translation performance of the target NMT models on WMT14 dataset.}
	\label{tab:dataset}

\end{table}

\begin{table}[!t]
	\centering
		\renewcommand{\arraystretch}{0.9}
	\setlength{\tabcolsep}{2.3pt}
	
	\scalebox{0.92}{
		\begin{tabular}[t]{@{} ccccccc @{}}
			\toprule[1pt]
		    \multirow{2}{*}{\textbf{Task}}  &
		    \multirow{2}{*}{\textbf{Method}}  & \multicolumn{5}{c}{\textbf{mBART50 (En-Fr)}} \\
			\cline{3-7}
			\rule{0pt}{2.5ex}    
			& & ASR$\uparrow$ & BLEU$\downarrow$ & chrF$\downarrow$ & Sim.$\uparrow$ & Perp.$\downarrow$ \\
			\midrule[1pt]
			\multirow{5}{*}{\rotatebox[origin=c]{90}{SST-2}} 
			& kNN &   9.22 & 34.57 & 60.14 & 0.77 & 6.64 \\
			& Seq2Sick & 1.48 & 61.20 & 70.93 & 0.74 & 0.72  
                \\
                & TransFool & 9.48 & 37.18 & 58.00 & 0.81 & 1.76 \\
                & ACT\textsubscript{TF} & 39.35 & \textbf{25.37} & \textbf{49.46} & \textbf{0.86} & 1.56 \\
                & ACT\textsubscript{BAE} & \textbf{41.04} &  38.00 & 57.84 & \textbf{0.86} & \textbf{0.61} \\
			\midrule[1pt]
			\multirow{5}{*}{\rotatebox[origin=c]{90}{MR}} 
			& kNN & 11.98 & 32.13 & 57.18 & 0.84 & 5.87 \\
			& Seq2Sick &  1.96 & 43.48 & 62.01 & 0.75 & 2.25 \\
                & TransFool &  12.71 & 39.17 & 61.92 & 0.83 & 2.26 \\ 
                & ACT\textsubscript{TF} & \textbf{40.71} & \textbf{12.13} & \textbf{36.88} & 0.82 & 2.56 \\
                & ACT\textsubscript{BAE} & 32.15 & 30.84 & 53.04 & \textbf{0.86} & \textbf{0.66} \\
			\midrule[1pt]
			\multirow{4}{*}{\rotatebox[origin=c]{90}{\scalebox{0.9}{AG's News}}} 
			& kNN & 2.99 & 49.17 & 68.53 & \textbf{0.96} & 1.37 \\
                & TransFool &  5.34 & 47.96 & 63.95 & 0.88 & 2.56 \\ 
                & ACT\textsubscript{TF} & \textbf{27.85} & \textbf{23.02} & \textbf{43.59} & 0.88 & 3.86 \\
                & ACT\textsubscript{BAE} & 9.07 & 32.11 & 52.41 & 0.95 & \textbf{1.15} \\
			
			\bottomrule[1pt]
		\end{tabular}
  }
  \caption{Evaluation results of the adversarial attacks against mBART50  model (En-Fr). 
	} \label{tab:mbart}
	
	\vspace{-5pt}
\end{table}

\begin{table*}[t]
	\centering
		\renewcommand{\arraystretch}{0.7}
	\setlength{\tabcolsep}{2pt}
 \scalebox{0.86}{

		\begin{tabular}[t]{@{} lcccccccccccc @{}}
			\toprule[1pt]
		    \multirow{2}{*}{\textbf{Dataset}}  & 
                \multirow{2}{*}{\textbf{\#Classes}} &
		     \multirow{2}{*}{\textbf{\#Train }} & 
		     \multirow{2}{*}{\textbf{\#Test }} &
		     \multirow{2}{*}{\textbf{Avg Length}} &
		    \multicolumn{2}{c}{\textbf{Marian NMT (En-Fr)}} &&
                \multicolumn{2}{c}{\textbf{Marian NMT (En-De)}} &&
		    \multicolumn{2}{c}{\textbf{mBART50 (En-Fr)}} \\
		    
		    \cline{6-7}
			\cline{9-10}
                \cline{12-13}   
			\rule{0pt}{2.5ex}    
			& &&&&  GPT-2 & BERT && GPT-2 & BERT && GPT-2 & BERT  \\
			
			\midrule[1pt]
			SST-2 &  2 & 67.3K & 0.9K & 17 & 88.88 & 90.94 && 86.01 & 84.40 && 86.01 & 88.30\\
			\midrule
			MR  & 2 & 8.5K  & 1K  & 20 & 81.33 & 84.43 && 79.55 & 80.11 && 78.61 & 82.08\\ 
			\midrule
			AG  & 4  & 120K & 7.6K & 43 & 93.47 & 93.53 && 93.86 & 92.78 && 93.46 & 93.55\\
			
			\bottomrule[1pt]
		\end{tabular}
  }
    \caption{Some statistics of the evaluation datasets, and the accuracy of the classifiers on the test sets.}
	\label{tab:data}
\end{table*}

\begin{figure*}[t]
\centering
\includegraphics[width=0.95\textwidth]{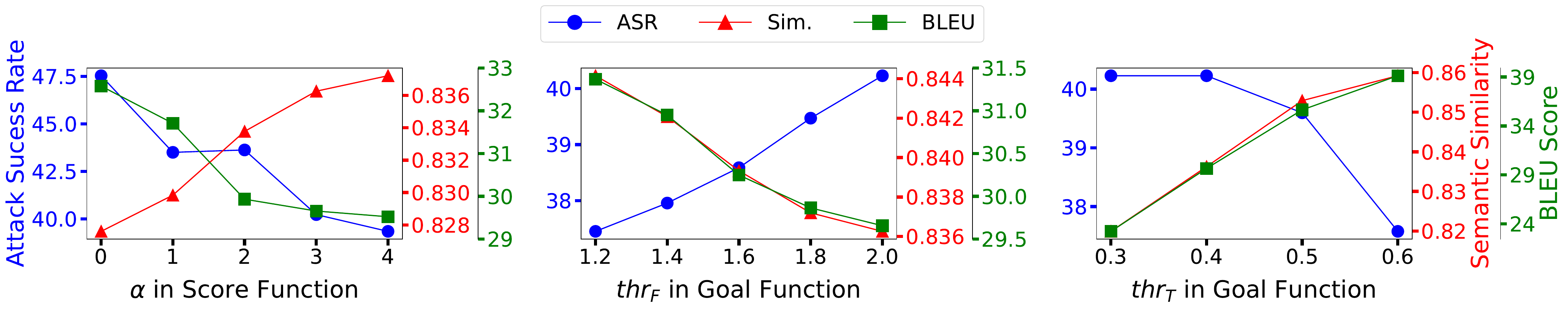} 
\caption{Effect of different parameters on  ACT\textsubscript{TF}  when attacking Marian NMT (En-Fr) on SST-2 dataset.}
\label{fig:abl}
\end{figure*}

\subsection{Effect of Parameters} \label{param}
Our attack has three parameters: the coefficient $\alpha$ in the score function, which controls the importance of translation in the word  ranking; the threshold $thr_T$ in the translation goal function; and the threshold $thr_F$ in the classification goal function. Figure \ref{fig:abl} demonstrates the effect of these parameters on the performance of ACT\textsubscript{TF} in terms of success rate, BLEU score, and semantic similarity. By increasing the coefficient $\alpha$, we are assigning more importance to the words that affect the translation, and hence, the BLEU score between the original and adversarial translations decreases.  
Moreover, by decreasing the threshold for the similarity between the original and adversarial translations $thr_T$ or by increasing the threshold for the logit difference of the classifier $thr_F$, the attack generates adversarial examples that are more successful  in changing the class of the adversarial translation, and they also have more impact on the translation.

\subsection{Ensemble of Classifiers} \label{ensemble}

\begin{table}[t]

	\centering
		\renewcommand{\arraystretch}{1}
	\setlength{\tabcolsep}{2.5pt}

		\begin{tabular}[t]{@{} lccccc @{}}
			\toprule[1pt]
		    \multirow{1}{*}{\textbf{Method}}  &
		    ASR$\uparrow$ & BLEU$\downarrow$ & chrF$\downarrow$ & Sim.$\uparrow$ & Perp.$\downarrow$ \\
			\midrule[1pt]
			\multirow{1}{*}{1 classifier} &  40.23 & 29.66 & 56.11 & 0.84 & 2.00  \\ 
			\multirow{1}{*}{2 classifiers} &   46.03 & 27.12 & 54.10 & 0.83 & 2.21 \\

			\bottomrule[1pt]
		\end{tabular}
  \caption{Performance of ACT\textsubscript{TF}  with 2 classifiers against Marian NMT (En-Fr)  on SST-2 dataset.} \label{tab:2class}

	\vspace{-12pt}
\end{table} 

In order to ensure that the classifier accurately predicts the class and that the attack targets the NMT model, an ensemble of classifiers can be used to find the class of the translation. This approach increases the reliability of the prediction made by the classifier. Table \ref{tab:2class} shows the attack performance when we use two classifiers in the attack process. These results show that we can increase the success rate and impact the translation more by using an ensemble of classifiers.

\begin{table}[t]
	\centering
		\renewcommand{\arraystretch}{1}
	\setlength{\tabcolsep}{2.5pt}

		\begin{tabular}[t]{@{} lccccc @{}}
			\toprule[1pt]
		    \multirow{1}{*}{\textbf{Method}}  &
		    ASR$\uparrow$ & BLEU$\downarrow$ & chrF$\downarrow$ & Sim.$\uparrow$ & Perp.$\downarrow$ \\
			\midrule[1pt]
			\multirow{1}{*}{BLEU} &  40.23 & 29.66 & 56.11 & 0.84 & 2.00  \\ 
			\multirow{1}{*}{BLEURT-20} &   45.39 & 33.98 & 55.79 & 0.82 & 2.56 \\

			\bottomrule[1pt]
		\end{tabular}
  \caption{Performance of ACT\textsubscript{TF}  with BLEURT-20 as the similarity metric against Marian (En-Fr)  on SST-2 dataset.} \label{tab:bleurt}

\end{table} 



\subsection{Influence of Similarity Metric for Translation} \label{sim}
In the previous experiments, we used BLEU score to measure the similarity between the original and adversarial translations. It has been shown that BLEURT-20 \cite{sellam2020learning} highly correlates with human judgments. However, the computation of this metric is time-consuming and makes the attack slow. We study the effect of the similarity metric used in our attack by using BLEURT-20 instead of BLEU score in our attack to Marian NMT (En-Fr) over SST-2 dataset. The results reported in Table \ref{tab:bleurt} show that the performance of our attack is consistent with our previous results when we use BLEURT-20. The success rate is indeed better in this case, but the run time increases to 83.18 seconds per sentence.

\subsection{Influence of the choice of classifier} \label{class}
In our attack framework, the adversary uses a classifier \textit{of its own} to find and change the class of output translation by the target NMT model. We study the choice of the classifier on the attack performance in Table \ref{tab:class}. In all our previous experiments, we fine-tuned a Language model with GPT-2 architecture on the training set of the attack's dataset, which is denoted by classifier 1. However, the attack may use an off-the-shelf classifier for the attack.  For the new attack, we use a  French sentiment classifier from HuggingFace, which is denoted by classifier 2.\footnote{The classifier is available at \url{https://huggingface.co/moussaKam/barthez-sentiment-classification}.} We should note that the accuracy of classifier 1 is 88.88, while classifier 2 has an accuracy of 83.72. The results show that with a less accurate classifier, the success rate slightly decreases.

\begin{table}[t]
	\centering
		\renewcommand{\arraystretch}{1}
	\setlength{\tabcolsep}{3pt}

		\begin{tabular}[t]{@{} lccccc @{}}
			\toprule[1pt]
		    \multirow{1}{*}{\textbf{Classifier}}  &
		    ASR$\uparrow$ & BLEU$\downarrow$ & chrF$\downarrow$ & Sim.$\uparrow$ & Perp.$\downarrow$ \\
			\midrule[1pt]
			\multirow{1}{*}{Classifier 1} &  40.23 & 29.66 & 56.11 & 0.84 & 2.00  \\ 
			\multirow{1}{*}{Classifier 2} &   31.78 & 28.98 & 57.61 & 0.86 & 1.73 \\

			\bottomrule[1pt]
		\end{tabular}
  \caption{Performance of ACT\textsubscript{TF}  with two different classifiers against Marian NMT (En-Fr) on SST-2 dataset.} \label{tab:class}
\end{table}

 \begin{table}[t]
	\centering
		\renewcommand{\arraystretch}{1}
	\setlength{\tabcolsep}{2.3pt}
	
	\scalebox{0.95}{

		\begin{tabular}[t]{@{} lcccccc @{}}
			\toprule[1pt]
		    \multirow{1}{*}{\textbf{Task}}  &
		    \multirow{1}{*}{\textbf{Model}}  & 
		    ASR$\uparrow$ & BLEU$\downarrow$ & chrF$\downarrow$ & Sim.$\uparrow$ & Perp.$\downarrow$ \\
			\midrule[1pt]
			\multirow{3}{*}{SST-2} 
                & M1 & 40.23 & 29.66 & 56.11 & 0.84 & 2.00  \\ 
                & M3 & 25.17 & 16.08 & 35.72 & 0.83 & 1.00  \\ 
                & M2 & 31.91 & 17.37 & 36.71 & 0.84 & 1.14 \\  
                
                \midrule
                
                \multirow{3}{*}{MR} 
                & M1 & 36.37 & 20.15 & 48.46 & 0.82 & 3.03
\\
                & M2 & 26.93 & 6.90 & 26.27 & 0.81 & 1.90\\
                & M3 & 32.38 & 9.23 & 26.90 & 0.83 & 2.39\\

               \midrule
                
                \multirow{3}{*}{AG} 
                & M1 & 22.84 & 25.62 & 49.19 & 0.85 & 6.02\\ 
                & M2 & 13.88 & 18.14 & 38.91 & 0.84 & 5.62\\
                & M3 & 14.87 & 21.07 & 41.42 & 0.84 & 5.18\\

			\bottomrule[1pt]
		\end{tabular}
  }
  \caption{Transferabiliy of ACT\textsubscript{TF}  from Marian (En-Fr), M1, to mBART50 (En-Fr) and Marian (En-De), M2 and M3, respectively.} \label{tab:transfer}

\end{table}

\subsection{Classification in the Source Language} \label{class_source} 
In our  attack, we generate adversarial examples that preserves the class in the source language while they change the class of the output translation by the target NMT model. We should note that  both TextFooler and BAE consider semantic similarity constraints for replacing the words in the input sentence. Therefore, they are able to preserve the class in the source language to some extent. To show this, we use a classifier in the source language (English) and check if the class of the adversarial and the original sentences were the same. For the attack against Marian NMT (En-Fr) over SST-2 dataset, 72\% of the adversarial sentences have the same class as the input sentence.\footnote{We use the finetuned BERT in \url{https://huggingface.co/gchhablani/bert-base-cased-finetuned-sst2}. The original accuracy of this model is 92\%.} This shows that although the constraints do not preserve the class completely, still for the majority of sentences, the class remains the same in the source language.

As an extension, we can add a constraint to explicitly force the adversarial examples to have the same class as the original sentence. We conduct the experiment for Marian NMT (En-Fr) over SST-2 dataset. The success rate decreases to 33.92\% (instead of 40.23\%). However, in this case, the class of the adversarial and original sentences are the same for 100\% of the cases.

\begin{table}[t]
	\centering
		\renewcommand{\arraystretch}{1.1}
	\setlength{\tabcolsep}{2.8pt}

		\begin{tabular}[t]{@{} lccccc @{}}
			\toprule[1pt]
		    \multirow{1}{*}{\textbf{Attack}}  &
		    ASR$\uparrow$ & BLEU$\downarrow$ & chrF$\downarrow$ & Sim.$\uparrow$ & Perp.$\downarrow$ \\
			\midrule[1pt]
			\multirow{1}{*}{Untargeted} &  22.84 & 25.62 & 49.19 & 0.85 & 6.02  \\ 
			\multirow{1}{*}{Targeted} &   11.20 & 24.57 & 50.41 & 0.85 & 4.17 \\

			\bottomrule[1pt]
		\end{tabular}
  \caption{Performance of ACT\textsubscript{TF} in the targeted setting against Marian (En-Fr)  on AG's News dataset.} \label{tab:target}

\end{table} 

\subsection{Transferability} \label{transfer}
We examine the transferability of our adversarial attack. In other words, we study whether adversarial samples crafted for one target NMT model  can also fool another NMT model. Inspired by \cite{sadrizadeh2023transfool}, we also  analyze  cross-lingual transferability, where the target languages of the two NMT models are different. Table \ref{tab:transfer} shows the transferability performance.  We use Marian NMT (En-Fr), denoted by M1, as the reference model and evaluate the transferability to mBART50 (En-Fr) and Marian NMT (En-De), which are denoted by M2 and M3, respectively. The results show that the attack is moderately transferable. We can also see that the adversarial examples that have more effect on the translation, i.e., with lower values of  BLEU score and chrF, are more transferable.

\begin{table*}[t]
	\centering
		\renewcommand{\arraystretch}{1.1}

	\setlength{\tabcolsep}{2pt}
	
	\scalebox{0.95}{

		\begin{tabular}[t]{@{} c|l| >{\parfillskip=0pt}p{11.5cm} @{}}
			\toprule[1pt]
		    \textbf{Task} & 
		    \textbf{Sentence}  & 
      \textbf{Text}\\
		
			\midrule[1pt]

   \multirow{7}{*}{\rotatebox[origin=c]{90}{SST-2}}  & \multirow{2}{*}{Org.} &  
      the notion that bombing buildings is the funniest thing in the world goes entirely unexamined in this startlingly unfunny comedy . \hfill\mbox{}  \\
			 

			& \multirow{2}{*}{Org. Trans. \quad (Negative)} &  
   L'idée que bombarder des immeubles est la chose la plus drôle du monde est totalement inexaminée dans cette comédie étonnamment peu amusante.  \hfill\mbox{}  \\
			 
			\cline{2-3}
			\rule{0pt}{2.5ex}

			& \multirow{2}{*}{Adv. ACT\textsubscript{BAE}} &  
  the notion that bombing buildings is the funniest \textbf{\textcolor{red}{one}} in the world goes entirely unexamined in this startlingly unfunny comedy . \hfill\mbox{} \\
			
			
		&	\multirow{3}{*}{Adv. Trans. \quad (Positive)} &  
   L'idée que la bombardement d'immeubles est le plus fun dans le monde va tout à fait étudiée dans cette comédie étonnamment fun. {\textcolor{blue}{(\textit{The translation means, "The idea that bombing buildings is the most fun in the world is thoroughly explored in this surprisingly fun comedy.", which is the total opposite of the input.})}} 
   \hfill\mbox{} \\

			\cline{1-3}
			\rule{0pt}{2.5ex} 

   \multirow{8}{*}{\rotatebox[origin=c]{90}{MR}}  & \multirow{2}{*}{Org.} &  
      while it's nothing we haven't seen before from murphy , i spy is still fun and enjoyable and so aggressively silly that it's more than a worthwhile effort . \hfill\mbox{}  \\
			 

			& \multirow{2}{*}{Org. Trans. \quad (Positive)} &  
Il n'y a rien que nous n'avons pas vu auparavant de murphy, j'espion est encore amusant et agréable et si agressivement stupide que c'est plus qu'un effort valable.
\hfill\mbox{}  \\
			 
			\cline{2-3}
			\rule{0pt}{2.5ex}

			& \multirow{2}{*}{Adv. ACT\textsubscript{BAE}} &  
  while it's \textbf{\textcolor{red}{material}} we haven't seen before from murphy , i spy is still \textbf{\textcolor{red}{interesting}} and enjoyable and so aggressively silly that it's more than a worthwhile effort .
 !
.   \hfill\mbox{} \\
			
			
		&	\multirow{2}{*}{Adv. Trans. \quad (Negative)} &  Il n'y a pas d'autre chose à faire, mais il n'y a pas d'autre chose à faire.
{\textcolor{blue}{(\textit{The translation is totally wrong, and it means, "There is nothing else to do, but there is nothing else to do."})}} 
   \hfill\mbox{} \\

			\bottomrule[1pt]

		\end{tabular}
	}
	\caption{Adversarial examples against mBART50 (En-Fr) in different Tasks.}
	\label{tab:sample1}
	
\end{table*}

\begin{table*}[!t]
	\centering
		\renewcommand{\arraystretch}{1.1}

	\setlength{\tabcolsep}{2pt}
	
	\scalebox{0.95}{

		\begin{tabular}[t]{@{} c|l| >{\parfillskip=0pt}p{11.5cm} @{}}
			\toprule[1pt]
		    \textbf{Task} & 
		    \textbf{Sentence}  & 
      \textbf{Text}\\
		
			\midrule[1pt]

   \multirow{4}{*}{\rotatebox[origin=c]{90}{SST-2}}  & Org. &  
      one of the more irritating cartoons you will see this , or any , year . 
  \hfill\mbox{}  \\
			 

			& Org. Trans. \quad (Negative) &  
   eine der irritierenden Karikaturen werden Sie dieses oder jedes Jahr sehen.
 \hfill\mbox{}  \\
			 
			\cline{2-3}
			\rule{0pt}{2.5ex}

			& Adv. ACT\textsubscript{TF} &  
  one of the more \textbf{\textcolor{red}{distasteful}} cartoons you will see this , or any , year . 
   \hfill\mbox{} \\
			
			
		&	\multirow{2}{*}{Adv. Trans. \quad (Positive)} &  
   einer der geschmackvollsten Karikaturen, die Sie sehen werden, dies, oder irgendein, Jahr.
 {\textcolor{blue}{(\textit{geschmackvollsten means "tastiest", which is a positive adjective.})}} 
   \hfill\mbox{} \\

			\cline{1-3}
			\rule{0pt}{2.5ex} 

   \multirow{4}{*}{\rotatebox[origin=c]{90}{MR}}  & Org. &  
      goofy , nutty , consistently funny . and educational !
 \hfill\mbox{}  \\
			 

			& Org. Trans. \quad (Positive) &  
   Goofy, nussig, durchweg lustig. und lehrreich!
\hfill\mbox{}  \\
			 
			\cline{2-3}
			\rule{0pt}{2.5ex}

			& Adv. ACT\textsubscript{TF} &  
   goofy , \textbf{\textcolor{red}{silly}} , \textbf{\textcolor{red}{ever comedic}} . and \textbf{\textcolor{red}{pedagogical}} !
.   \hfill\mbox{} \\
			
			
		&	\multirow{1}{*}{Adv. Trans. \quad (Negative)} &  Dumme, dumme, immer komische und pädagogische!
{\textcolor{blue}{(\textit{dumme means "stupid" and is repeated twice, making the sentence negative.})}} 
   \hfill\mbox{} \\

			\bottomrule[1pt]

		\end{tabular}
	}
	\caption{Adversarial examples against Marian NMT (En-De) in different Tasks.}
	\label{tab:sample3}
	
\end{table*}

\subsection{Targeted attack} \label{target}
We can extend our attack to the targeted settings, where the adversary aims to change the translation such that it belongs to a specific class. To this end, we can change the goal function of equation \eqref{logit} as:

\begin{equation}\label{logit_target}
    \mathbf{w}'_t - \max\limits_{i\neq t}{\mathbf{w}'_i} > thr_F,
\end{equation}
where $\mathbf{w'} = W(\mathbf{y'})$ are the logits, and $t$ is the predefined target class. This ensures that the class of adversarial translation is predicted as the target class by the classifier  with high confidence. We evaluate ACT\textsubscript{TF} in this setting against Marian NMT (En-Fr) on AG's News dataset when the target class is "World". The results in Table \ref{tab:target} show that although this setting is more challenging than the untargeted setting, our attack is still successful.

\begin{table*}[!t]
	\centering
		\renewcommand{\arraystretch}{1.1}

	\setlength{\tabcolsep}{2pt}
	
	\scalebox{0.95}{

		\begin{tabular}[t]{@{} c|l| >{\parfillskip=0pt}p{11.5cm} @{}}
			\toprule[1pt]
		    \textbf{Task} & 
		    \textbf{Sentence}  & 
      \textbf{Text}\\
		
			\midrule[1pt]

   \multirow{7}{*}{\rotatebox[origin=c]{90}{SST-2}}  & Org. &  
      paid in full is so stale , in fact , that its most vibrant scene is one that uses clips from brian de palma 's scarface .  \hfill\mbox{}  \\
			 

			& Org. Trans. \quad (Negative) &  
   payé dans son intégralité est tellement sombre, en fait, que sa scène la plus dynamique est celle qui utilise des clips de la cicatrice de Brian de palma. \hfill\mbox{}  \\
			 
			\cline{2-3}
			\rule{0pt}{2.5ex}

			& Adv. ACT\textsubscript{TF} &  
   paid in full is so stale , in \textbf{\textcolor{red}{circumstance}} , that its most vibrant scene is one that \textbf{\textcolor{red}{used}} clips from brian de palma 's scarface .   \hfill\mbox{} \\
			
			
		&	\multirow{1}{*}{Adv. Trans. \quad (Positive)} &  
   payé en totalité est si stable, dans les circonstances, que sa scène la plus dynamique est celui qui a utilisé des clips de Brian de palma's cicatrice. {\textcolor{blue}{(\textit{Stale is translated as stable, which is a positive adjective.})}} 
   \hfill\mbox{} \\

			\cline{1-3}
			\rule{0pt}{2.5ex} 

   \multirow{11}{*}{\rotatebox[origin=c]{90}{AG}}  & Org. &  
      Charges reduced for Iraq jail MP MANNHEIM, Germany -- A US military policewoman accused in the Abu Ghraib prison abuse scandal had the charges against her reduced yesterday as a set of pretrial hearings wrapped up at an American base in Germany. \hfill\mbox{}  \\
			 

			& Org. Trans. \quad (World) &  
   Des accusations réduites pour la prison irakienne MP MANNHEIM, Allemagne -- Une policière militaire américaine accusée dans le scandale d'abus de prison d'Abu Ghraib a eu les accusations réduites contre elle hier comme un ensemble d'audiences préliminaires terminées dans une base américaine en Allemagne. \hfill\mbox{}  \\
			 
			\cline{2-3}
			\rule{0pt}{2.5ex}

			& Adv. ACT\textsubscript{TF} &  
   \textbf{\textcolor{red}{Charging}} reduced for Iraq jail MP \textbf{\textcolor{red}{GRAZ}}, Germany -- A US military policewoman accused in the Abu Ghraib prison abuse scandal \textbf{\textcolor{red}{am}} the charges against her reduced yesterday as a games of pretrial hearings wrapped up at an American base in \textbf{\textcolor{red}{Deutsche}}.   \hfill\mbox{} \\
			
			
		&	\multirow{1}{*}{Adv. Trans. \quad (Sport)} &  
   Les accusations portées contre elle ont été réduites hier, alors qu'un jeu d'audiences préliminaires s'est déroulé dans une base américaine en Deutsche. {\textcolor{blue}{(\textit{The first part of the sentence, which has critical information, is not translated at all.})}} 
   \hfill\mbox{} \\

                \cline{1-3}
			\rule{0pt}{2.5ex} 
			
		    \multirow{4}{*}{\rotatebox[origin=c]{90}{MR}}  & Org. &  
      williams absolutely nails sy's queasy infatuation and overall strangeness . \hfill\mbox{}  \\
			 

			& Org. Trans. \quad (Positive) &  
   williams absolument clous l'engouement de Sy et l'étrangeté globale. \hfill\mbox{}  \\
			 
			\cline{2-3}
			\rule{0pt}{2.5ex}

			& Adv. ACT\textsubscript{BAE} &  
   williams \textbf{\textcolor{red}{too}} nails sy's queasy \textbf{\textcolor{red}{demeanor}} and \textbf{\textcolor{red}{other}} strangeness . \hfill\mbox{} \\
			
			
		&	\multirow{1}{*}{Adv. Trans. \quad (Negative)} &  
   waliams trop clous sy de comportement bizarre et autre étrangeté. {\textcolor{blue}{(\textit{trop conveys a negative sentiment.})}} 
   \hfill\mbox{} \\

                \cline{1-3}
			\rule{0pt}{2.5ex} 
			
	\multirow{11}{*}{\rotatebox[origin=c]{90}{AG}} &	    \multirow{2}{*}{Org.} &  
      Airlines Agree to Cuts at O'Hare Federal officials today announced plans to temporarily cut 37 flights operating at Chicago's O'Hare International Airport to help reduce the delay problems that ripple across the country. \hfill\mbox{}  \\
			 

		&	\multirow{1}{*}{Org. Trans. \quad (Business)} &  
 Les compagnies aériennes conviennent de réduire les vols à O'Hare Les responsables fédéraux ont annoncé aujourd'hui leur intention de réduire temporairement 37 vols à l'aéroport international O'Hare de Chicago afin de réduire les problèmes de retard qui se posent à travers le pays. \hfill\mbox{}  \\
			 
			\cline{2-3}
			\rule{0pt}{2.5ex}

		&	\multirow{2}{*}{Adv. ACT\textsubscript{BAE}} &  
     Airlines Agree to Cuts at \textbf{\textcolor{red}{airports}} Federal officials today announced plans to temporarily cut 37 \textbf{\textcolor{red}{boeing}} operating at Chicago's O'Hare International Airport to help reduce the \textbf{\textcolor{red}{continuing}} problems that ripple \textbf{\textcolor{red}{by}} the country.   \hfill\mbox{} \\
			
			
		&	\multirow{2}{*}{Adv. Trans. \quad (World)} &  
   Les compagnies aériennes s'engagent à réduire les émissions dans les aéroports Les responsables fédéraux ont annoncé aujourd'hui leur intention de couper temporairement 37 sangliers à l'aéroport international O'Hare de Chicago pour aider à réduire les problèmes persistants que connaît le pays.
   {\textcolor{blue}{(\textit{Boeing is translated as sangliers, which means "boar", and hence, the category of the translation is changed from Business to World.})}} \hfill\mbox{} \\

			\bottomrule[1pt]
			

		\end{tabular}
	}
	\caption{Adversarial examples against Marian NMT (En-Fr) in different Tasks.}
	\label{tab:sample2}
	
\end{table*}

\subsection{More Adversarial Examples} \label{example}

In Tables \ref{tab:sample1}-\ref{tab:sample3}, we present more adversarial examples generated by ACT\textsubscript{TF} and ACT\textsubscript{BAE} against various NMT models. These examples highlight how our proposed attack can generate adversarial examples whose translations have different classes than the original translations, which can be harmful for the users.

\section{Discussion}\label{discussion}
The recent parallel work of \cite{raina2023sentiment} proposes an attack against NMT models to change the human perception of the translation, specifically sentiment. They use a sentiment classifier to approximate human perception. In our attack framework, the adversary aims to mislead the NMT model such that the class of the adversarial translation differs from that of the original translation. Therefore, the attacker uses a classification model to guide the attack. Based on the attack objective, the adversary can use a classifier suitable for any task and not just a sentiment classifier. By focusing on the class of the output translation, the adversarial attack has more impact on the translation since the class of the translation reflects the overall meaning. 

Evaluating adversarial attacks against NMT models is challenging since the perturbation to the input may directly appear in the translation and change the ground-truth output. By using the classification objective, we can provide a more comprehensive assessment of the attack's impact on the NMT model.
To evaluate the robustness of NMT models to our attack, we introduce modifications to existing black-box word-replacement attacks. Since the target of the attack is the NMT model, we propose a new goal function to distinguish between the impact of the attack on the NMT model and the classifier. However, the attack proposed by  \cite{raina2023sentiment} does not have a mechanism to ensure that it specifically misleads the NMT model rather than the classifier. We also study different scenarios that can be considered in our framework, e.g., when the target of the attack is the entire system (NMT + classifier), or just the classifier, which are not studied in \cite{raina2023sentiment}. Finally, we extensively evaluate  the robustness of NMT models to our attack framework by considering different tasks and NMT models, various performance metrics, and a comparison to baselines. In contrast, their experiments appear to be  limited.

\end{document}